\documentclass[a4paper]{llncs}

 \usepackage{svg}
 \usepackage{csquotes}
 \usepackage{hyperref}
 \usepackage{graphicx}
 \usepackage{mathtools}

 \usepackage{amssymb}
 \usepackage{colortbl}
 \usepackage{multirow}
 \usepackage{footnote}
 \usepackage{algorithm}
\usepackage[noend]{algpseudocode}
 \usepackage{cite}
 \usepackage{bbding}
 
 \makesavenoteenv{tabular}
 \makesavenoteenv{table}

\begin{document}
 	
 	\title{Towards Model-based Reinforcement Learning for Industry-near Environments}
 	\titlerunning{Reinforcement Learning for Industry-near Environments}

	\author{Per-Arne Andersen\textsuperscript{\Envelope}\orcidID{0000-0002-7742-4907} 
	\and Morten Goodwin\orcidID{0000-0001-6331-702X}
	\and Ole-Christoffer Granmo\orcidID{0000-0002-7287-030X}}
 	

 	%
 	\authorrunning{P. Andersen et al.}
 	\institute{Department of ICT, University of Agder, Grimstad, Norway\\
 		\email{\{per.andersen,morten.goodwin,ole.granmo\}@uia.no}}
 	\maketitle              
 	\begin{abstract}
		Deep reinforcement learning has over the past few years shown great potential in learning near-optimal control in complex simulated environments with little visible information. Rainbow (Q-Learning) and PPO (Policy Optimisation) have shown outstanding performance in a variety of tasks, including Atari 2600, MuJoCo, and Roboschool test suite. While these algorithms are fundamentally different, both suffer from high variance, low sample efficiency, and hyperparameter sensitivity that in practice, make these algorithms a no-go for critical operations in the industry.
		
		On the other hand, model-based reinforcement learning focuses on learning the transition dynamics between states in an environment. If these environment dynamics are adequately learned, a model-based approach is perhaps the most sample efficient method for learning agents to act in an environment optimally. The traits of model-based reinforcement are ideal for real-world environments where sampling is slow and for mission-critical operations. In the warehouse industry, there is an increasing motivation to minimise time and to maximise production. Currently, autonomous agents act suboptimally using handcrafted policies for significant portions of the state-space.
		
		In this paper, we present The Dreaming Variational Autoencoder v2 (DVAE-2), a model-based reinforcement learning algorithm that increases sample efficiency, hence enable algorithms with low sample efficiency function better in real-world environments. We introduce Deep Warehouse, a simulated environment for industry-near testing of autonomous agents in grid-based warehouses. Finally, we illustrate that DVAE-2 improves the sample efficiency for the Deep Warehouse compared to model-free methods.
		
 	\end{abstract}
 
 \pagestyle{plain} 
 \setcounter{page}{1}
 \section{Introduction}
The goal of reinforcement learning is to maximise some notion of feedback through interaction with an environment \cite{Sutton2018}. The environment can be known, which makes this learning process trivial, or have hidden state information, which typically increases the complexity of learning significantly. In model-free reinforcement learning, actions are sampled from some policy that is optimised indirectly through direct policy search (Policy gradients), a state-value function (Q-learning), or a combination of these (Actor-Critic). There are many recent contributions to these algorithms that increase sample efficiency \cite{Buckman2018}, reduce variance \cite{greensmith2004variance}, and increase training stability \cite{Schulman2017}.

It is challenging to deploy model-free methods in real-world environments because current state-of-the-art algorithms require millions of samples before any optimal policy is learned. Due to this, model-based reinforcement learning is an appealing approach because it has significantly better sample efficiency compared to the model-free methods \cite{Kaelbling1996}. The goal of model-based algorithms is to learn a predictive model of the real environment that is used to learn the controller of an agent. The downside of model-based reinforcement learning is that the predictive model may become inaccurate for longer time-horizons, or collapse entirely in areas of state-space that has not observed.

We propose a model-based reinforcement learning approach for industry-near systems where a predictive model is learned without direct interaction with the environment. We use Automated Storage and Retrieval Systems (ASRS) to benchmark our proposed algorithm. Learning a predictive model of the environment is isolated from the physical environment, which guarantees safety during training. If a predictive model is sufficiently trained, a model-free algorithm, such as DQN \cite{Mnih2013} can be trained off-line. Training can be done in a large-scale distributed setting, which significantly reduces the training time. When the model-free algorithm is trained sufficiently, it will be able to replace a sub-optimal expert-system with minimal effort.

The paper is organised as follows. Section~\ref{sec:literature_review} discusses the current state of the art in model-based reinforcement learning, and familiarise the reader of recent work in ASRS systems. Section~\ref{sec:bg} briefly outlines relevant background literature on reinforcement learning. Section~\ref{sec:dvae} introduces the DVAE-2 algorithm and details the architecture thoroughly. Section~\ref{sec:deep_warehouse} proposes the Deep Warehouse, a novel high-performance environment for industry-near testing of reinforcement learning algorithms. Section~\ref{sec:experiments} presents our results using DVAE-2 in various environments, including complex environments such as Deep Warehouse, Deep RTS and Deep Line Wars. Finally, section~\ref{sec:conclusion} concludes our work and outlines a roadmap for our future work.

 \section{Literature Review}
 \label{sec:literature_review}
Reinforcement Learning is a maturing field in artificial intelligence, where a significant portion of the research is concerned with model-free approaches in virtual environments. Reinforcement learning methods in large-scale industry-near environments are virtually absent from the literature. The reason for this could be that (1) model-free methods do not give the sample efficiency required and that (2) there is little evidence that model-based approaches achieve reliable performance. In this section, we briefly discuss the previous work in ASRS systems and present promising results for model-based reinforcement learning.

\subsection{Automated Storage and Retrieval Systems (ASRS)}
There is to our knowledge no published work where reinforcement learning schemes are used to control taxi-agents in ASRS environments. The literature is focused on heuristic-based approaches, such as tree-search and traditional pathfinding algorithms. In \cite{Roodbergen2009}, a detailed survey of the advancements in ASRS systems which categorise an ASRS system into five components; System Configuration, Storage Assignment, Batching, Sequencing, and Dwell-point. We adopt these categories in search of a reinforcement learning approach for ASRS systems 

\subsection{Model-based Reinforcement Learning}
In model-based reinforcement learning, the goal is to learn state-transitions based on observations from the environment, the predictive model. If the predictive model is stable, with low variance and improves monotonically during training, it is, to some degree, possible to learn model-free agents to act optimally in environments that have never been observed directly.

Perhaps the most sophisticated algorithm for model-based reinforcement learning is the Model-based policy optimisation (MBPO) algorithm, proposed by Janner et al. \cite{Janner2019} The authors empirically show that MBPO performs significantly better in continuous control tasks compared to previous methods. MBPO proves to be monotonically improving given that the following bounds hold:
\begin{align*}
\eta \lbrack \pi \rbrack \geq \hat{\eta}  \lbrack \pi \rbrack - C
\end{align*}
where \(\eta \lbrack \pi \rbrack\) denotes the returns in the real environment under a policy whereas \(\hat{\eta}  \lbrack \pi \rbrack\) denotes the returns in the predicted model under policy \(\pi\). Furthermore, the authors show that as long as they can improve the C, the performance will increase monotonically \cite{Janner2019}.

Gregor et al. proposed a scheme to train expressive generative models to learn belief-states of complex 3D environments with little prior knowledge. Their method was effective in predicting multiple steps into the future (overshooting) and significantly improve sample efficiency. In the experiments, the authors illustrated model-free policy training in several environments, including DeepMind Lab. However, the authors found it difficult to use their predictive model in model-free agents directly. \cite{Gregor2019}


Neural Differential Information Gain Optimisation (NDIGO) algorithm by Azar et al. is a self-supervised exploration model that learns a world model representation from noisy data. The primary features of NDIGO are its robustness to noise due to their method to cancel out negative loss and to give positive learning more value. The authors show in their maze environment that the model successfully converges towards an optimal world model even when introducing noise. The author claims that the algorithm outperforms previous state-of-the-art, being the Recurrent World Model from. \cite{Azar2019}

The Dreaming Variational Autoencoder (DVAE) is an end-to-end solution for prediction the probable future state \(p(\hat{s}_{t+1} | s_t, a_t \). The authors showed that the algorithm successfully predicted next state in non-continuous environments and could with some error predict future states in continuous state-space environments such as the Deep Line Wars environment. In the experiments, the authors used DQN, PPO, and TRPO using an artificial buffer to feed states to the algorithms. In all cases, the DVAE algorithm was able to create buffers that were accurate enough to learn a near-optimal policy. \cite{Andersen2018b}

The algorithm VMAV-C is a combination of VAE and attention-based value function (AVF), and mixture density network recurrent neural network (MDN-RNN) from \cite{Ha2018a}. This modification to the original World Models algorithm improved performance in the Cart Pole environment.  They used the on-policy algorithm PPO to learn the optimal policy from the latent representation of the state-space \cite{Liang2018}.

Deep Planning Network (PlaNet) is a model-based agent that interpret the pixels of a state to learn a predictive model of an environment. The environment dynamics are stored into latent-space, where the agent sample actions based on the learned representation. The proposed algorithm showed significantly better sample efficiency compared to model-free algorithms such as A3C~\cite{Hafner2018}.

In \textit{Recurrent World Models Facilitate Policy Evolution}, a novel architecture for training RL algorithms using variational autoencoders. This paper showed that agents could successfully learn the environment dynamics and use this as an exploration technique requiring no interaction with the target domain. The architecture is mainly three components; vision, controller, and model, the vision model is a variational autoencoder that outputs a latent-space variable of an observation. The latent-space variable is processed in the model and is fed into the controller for action decisions. Their algorithms show state-of-the-art performance in self-supervised generative modelling for reinforcement learning agents. \cite{Ha2018a}

Chua et al. proposed \textit{Probabilistic Ensembles with Trajectory Sampling} (PETS). The algorithm uses an ensemble of bootstrap neural networks to learn a dynamics model of the environment over future states. The algorithm then uses this model to predict the best action for future states. The authors show that the algorithm significantly lowers sampling requirements for environments such as half-cheetah compared to SAC and PPO. \cite{Chua2018}

DARLA is an architecture for modelling the environment using $\beta$-VAE \cite{Higgins2016}. The trained model was used to learn the optimal policy of the environment using algorithms such as DQN \cite{Mnih2013}, A3C, and Episodic Control \cite{Blundell2016}. DARLA is to the best of our knowledge, the first algorithm to introduce learning without access to the ground-truth environment during training.

 \section{Background}
 \label{sec:bg}
 \newcommand{\rfn}[0]{$r: \mathcal{S} \times \mathcal{A} \rightarrow \mathbb{R}$}
 \newcommand{\mdp}{
	($\mathcal{S}$, 
	$\mathcal{A}$, 
	$r$, 
	$\mathcal{P}$, 
	$\mathcal{P}_0$, 
	$\gamma$)
 }
 
\footnotetext[1]{$\mathcal{S}$ and $\mathcal{A}$ is defined for discrete or continuous spaces. \rfn~where $r$ is commonly referred to as  $\mathcal{R}(s, s')$ in the literature.}

  \begin{figure}
  	\centering
 	\includegraphics[width=0.7\textwidth]{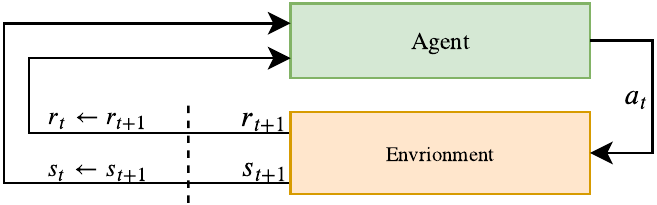}
 	\caption{The agent-environment interaction in a Markov decision process \cite{Sutton2018}}
 	\label{fig:mdp}
 \end{figure}
Markov decision processes (MDP's) are a mathematical framework commonly used to define reinforcement learning problems, as illustrated in Figure \ref{fig:mdp}. In an MDP, we consider the tuple \mdp \footnotemark  where $\mathcal{S}$ is  the  state space, $\mathcal{A}$ is the action space available to the agent, \rfn~is the expected immediate reward function, $P$ is the transition function which defines the probability $\mathcal{P}(s', s, a)  = \mathcal{P}r(s' | s, a)$ and $\mathcal{P}_0$ is the probability for the initial state $s_0$.
 
The goal of a reinforcement learning agent is to encourage good behaviour and to discourage bad behaviour. Optimal behaviour is achieved when the agent finds a composition of parameters that maximise its performance, thus finds the optimal policy$\pi^*$. Consider

 \begin{equation}
 	\pi^* = \arg \max_{\pi \in \Pi}J(\pi)\text{,}
 \end{equation}
where $J(\pi)$ is the objective function for maximising the expected discounted reward defined as

\begin{equation}
	J(\pi) = \mathbb{E}_{s_0, a_0, s_1,\dots} \Bigg \lbrack \sum_{t=0}^{\infty} \gamma^t r(s_t, a_t)~|~\pi, s_0 \sim \mathcal{P}_0 \Bigg \rbrack\text{,}
\end{equation}
 where $\gamma \in (0, 1)$ is the discounting factor of future rewards. If $\gamma = 1$, all future state rewards are accounted for equally, while $\gamma = 0$, we are only concerned about the current state.

\section{Learning policies using predictive models}
\label{sec:dvae}
The Dreaming Variational Autoencoder v2 (DVAE-2) is an architecture for learning a predictive model of arbitrary environments \cite{Andersen2018b}. In this work, we aim to improve the first version of the DVAE for better performance in real-world environments. A common problem in model-based reinforcement learning is that it takes millions of samples to generalise well across sparse data. We aim to approve sample efficiency from the original DVAE and if possible, surpass the performance of model-free methods.

\subsection{Motivation and Environment Safety}
\begin{figure}
	\includegraphics[width=\textwidth]{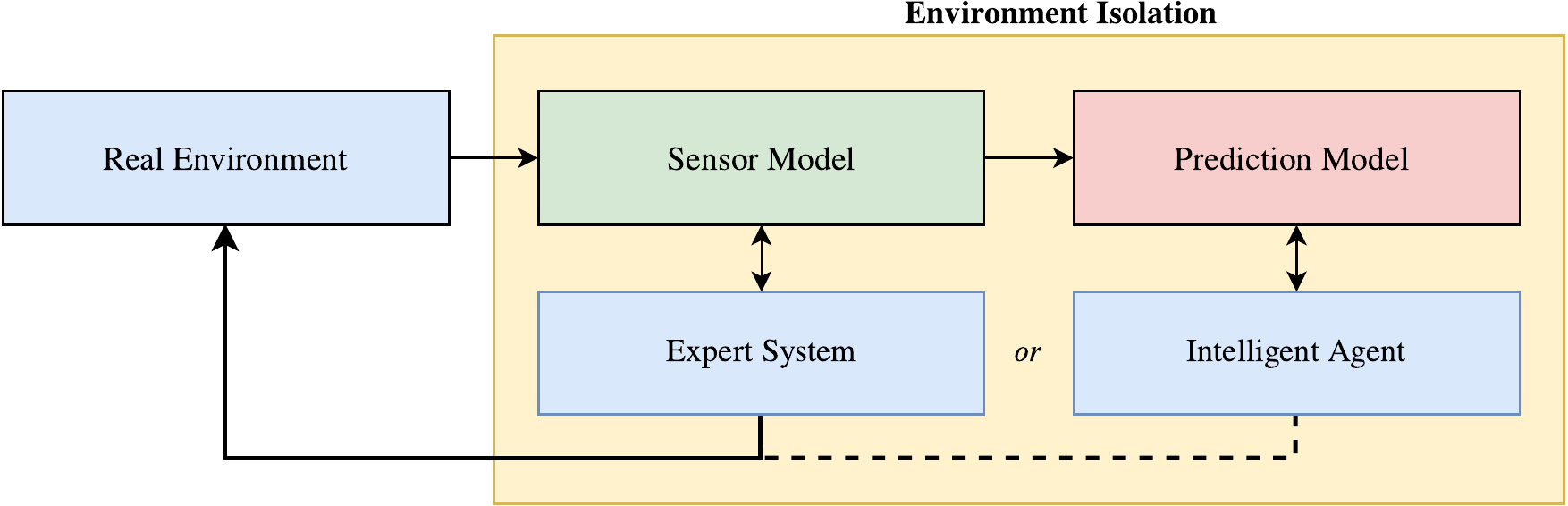}

	\caption{The proposed model isolates the intelligent agent from the mission-critical sensor model. The real environment projects onto a sensor model that the expert system uses to control taxis in a real environment. The predictive model observes the behaviour of the sensor model and the actions performed by the expert system. The predictive model is trained using error gradients, where the loss is the distance between the sensor model and the predictive model. When the error becomes sufficiently low, an intelligent agent can be trained using only data from the predictive model. Assuming that the intelligent agent converges to some performance threshold, it can be deployed as a drop-in replacement to the expert system.} \label{fig:overview_real_to_learned_intelligent_agent}
\end{figure}

Figure~\ref{fig:overview_real_to_learned_intelligent_agent} shows an abstract overview of DVAE-2 training in an environment. In real-world, industry-near environments, there is little room for interruptions. In model-free reinforcement learning, the agent interacts with the environment to learn its policy. Because this is not possible in many real-world environments, the DVAE-2 algorithm only observes during training. During training, the DVAE-2 algorithm learns how the transition function behaves and learns an estimated state-value function  \(V\) that represent the value of being in that current state.

\subsection{The Dreaming Variational Autoencoder v2}
\begin{figure}
	\includegraphics[width=\textwidth]{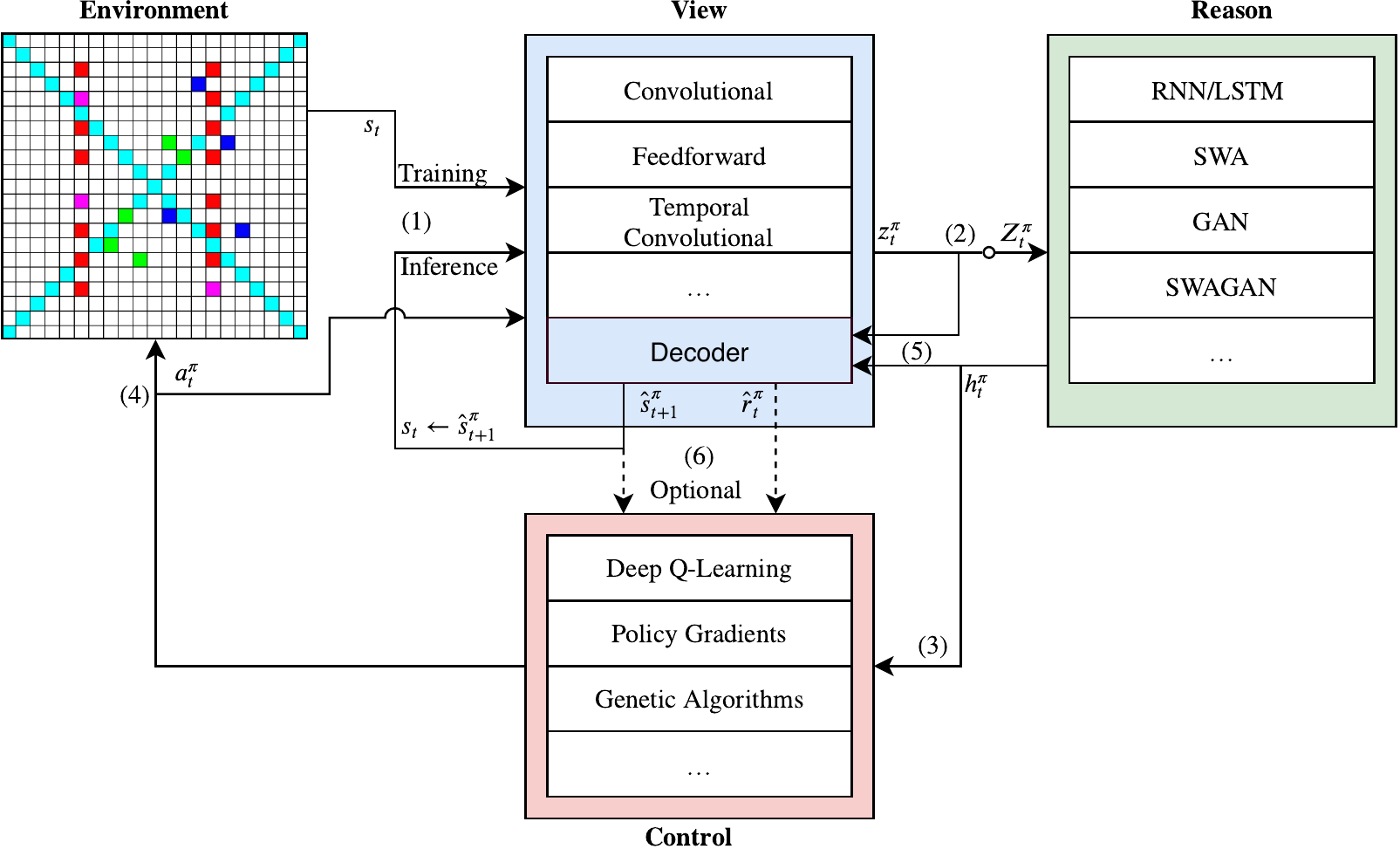}
	\caption{The component-based DVAE-2 architecture.}
	\label{fig:dvae_arch}
\end{figure}
The original DVAE architecture had severe challenges with modelling of continuous state-spaces \cite{Andersen2018b}, and many algorithms were added to the model to improve performance across various environments including autoencoders, LSTMs, and fine-tuned variations of these. The DVAE-2 extends this with a split into three individual components; forming the \textbf{V}iew, \textbf{R}eason and \textbf{C}ontrol (VRC) model. The VRC model embeds all improvements into a single model and learns which algorithms to use under certain conditions in an environment  

Figure \ref{fig:dvae_arch} shows an overview of the proposed VRC. (1) A state \(s_t\) is observed. During training, this observation stems from the real-environment while at inference time, from the predictive model. The observation  is encoded in the \textbf{view} component (e.g. via AE or GAN) and outputs an embedding \(z\) at time \(t\) w.r.t policy \(\pi\). (2) The \textbf{reason} component learns the time dynamics between state sequences. Encoded states are accumulated into a buffer \(Z_t^\pi = \lbrace z_{t-n} \dots z_t  \rbrace^\pi \) and are then used to predict the hidden-state  \(h_t^\pi\) w.r.t the encoded state sequence.  The reason component typically consists of a model with RNN-like structure that generalises well on sequence data. (3) The hidden state is then used to evaluate an action using policy \(\pi\), and (4) is sent to the environment and the view for the next iteration. (5) The decoder, prepares the hidden-state \(h_t^\pi\) and encoded state \(z_t^\pi\), producing the succeeding state \(\hat{s}_{t+1}^\pi\). The prediction is then used in the next iteration as current state \(s_t\), which leads back to (1). As an optional mechanism, the controller can use the output from the decoder, instead of the hidden state information. This is beneficial when working with model-free algorithms such as deep q-networks \cite{Mnih2013}.

\subsection{Model selection}
During technique selection in the components, we perform the following evaluation. An observation \(s_t\) is sent to the view component of DVAE-2. All of the view techniques are initially assumed to be uniformly qualified to encode and predict future states. For each iteration, the computed error is summarised as a score, and during inference, the technique with the lowest score is used\footnote{In this setting, the lowest score is the technique with least accumulated error.}. We use the same method for determining the best reasoning algorithm in a specific environment.

\subsection{Implementation}
The implementation of the DVAE-2 algorithm with dynamic component selection enabled several significant improvements to over the previous DVAE model\cite{Andersen2018b}. Notably, the k-step model rollout from \cite{Janner2019} is implemented to stabilise training. We found that using shorter model-rollouts provided better control policies, but at the cost of higher sample efficiency. Also, by embedding time into the encoded state improved the model stability and prediction capabilities \cite{Ha2018}. The  DVAE-2 algorithm is defined  as follows.
\begin{algorithm}[H]
	\caption{DVAE-2: Minimal Implementation}
	\label{alg:dvae} 
	\begin{algorithmic}[1]
		\State Initialize policy \(\pi_\theta(s_t|a_t)\), predictive model \(p_\psi(\hat{s}_{t+1}, \hat{r}, h_t | s_t, a_t^\pi)\)
		\State Let \(Z = \lbrace z_{t-n}^\pi \dots z_t^\pi \rbrace\), a vector of encoded states
		\State Initialize encoder \(ENC(z_t^\pi | s_t, a_t^\pi)\), temporal reasoner \(TR(h_t^\pi | Z)\) 
		
		\For {N epochs}
		\State \(\mathcal{D}_{env} \leftarrow\) Collect samples from \(p_{env}\) under predefined policy \(\pi\)
		\State Train model \(p_\psi\) on data batch \(D_{env}\) via \(MLE\)\footnotemark
		\EndFor
		
		\For {M epochs}
		\State Sample initial state  \(s_0 \sim U(0,1)\) from \(\mathcal{D}_{env}\)
		\State Construct \(\lbrace \mathcal{D}_{p_\psi} | t < k,  TR(h_t^{\pi_\theta}  | ENC(z_t | s_t, a_t)^{\pi_\theta}), s_t = s_0\rbrace\)
		\State Update policy \(\pi_\theta\) using pairs of \((\hat{s}_t, a_t, \hat{r}_t, \hat{s}_{t+1})^{\pi_\theta}\)
		\EndFor

	\end{algorithmic}
\end{algorithm}
\footnotetext{We use the mean squared error (MSE) loss in our implementation.}

Algorithm~\ref{alg:dvae} works as follows. (Line 1) We initialise the control policy and the predictive model (DVAE-2) parameters. (Line 2) The \(Z\) variable denotes a finite set of sequential view model (ENC) predictions that are used to capture time dependency between states in the reason model (TR). (Line 5) We collect samples from the real environment \(p_{env}\) under a predefined policy, such as an expert system, see Figure~\ref{fig:overview_real_to_learned_intelligent_agent}. (Line 6) The predictive model \(p_\psi\) is then trained using the collected data \(\mathcal{D}_{env}\) via maximum likelihood estimation. In our case, we use mean squared error to measure the error distance \(MSE(p_\psi \| p_{env})\). When the DVAE-2 algorithm has trained sufficiently, the model-free algorithm will train for \(M\) epochs (Line 7) using the predictive model \(p_\psi\) instead of \(p_{env}\). (Line 8) First, we sample the initial state \(s_0\) uniformly from the real dataset \(\mathcal{D}_{env}\). (Line 9) We then construct a prediction dataset \(\mathcal{D}_{p_{\psi}}\) and predict future states using the control policy (i.e. sampling from the predictive model). (Line 10) The parameterised control policy is then optimised using  \((\hat{s}_t, a_t, \hat{r}_t, \hat{s}_{t+1})^{\pi_\theta}\) pairs during rollouts.

\section{The Deep Warehouse Environment}
\label{sec:deep_warehouse}
Training algorithms in real-world environments is known to have severe safety challenges during training and suffers from low sampling speeds \cite{Botvinick2019}. It is therefore practical, to create a simulation of the real environment so that researches can quickly test algorithm variations with quick feedback on its performance.

This section presents the Deep Warehouse\footnote{The deep warehouse environment is open-source and freely available at \url{https://github.com/cair/deep-warehouse}} environment for discrete and continuous action and state spaces. The environment has a wide range of configurations for time and agent behaviour, giving it tolerable performance in simulating proprietary automated storage and retrieval systems.

\subsection{Motivation}
In the context of warehousing, an Automated Storage and Retrieval System (ASRS) is a composition of computer programs working together to maximise the incoming and outcoming throughput of goods. There are many benefits of using an ASRS system, including high scalability, increased efficiency, reduced operating expenses, and operation safety. We consider a cube-based ASRS environment where each cell is stacked with item containers. On the surface of the cube, taxi-agents are collecting and delivering goods to delivery points placed throughout the surface. The taxi-agents are controlled by a computer program that reads sensory data from the taxi and determines the next action.

Although these systems are far better than manual labour warehousing, there is still significant improvement potential in current state-of-the-art. Most ASRS systems are manually crafted expert systems, which due to the high complexity of the multi-agent ASRS systems only performs sub-optimally. \cite{Roodbergen2009}.

\subsection{Implementation}

\begin{figure}
	\centering
	\fbox{\includegraphics[width=0.5\textwidth]{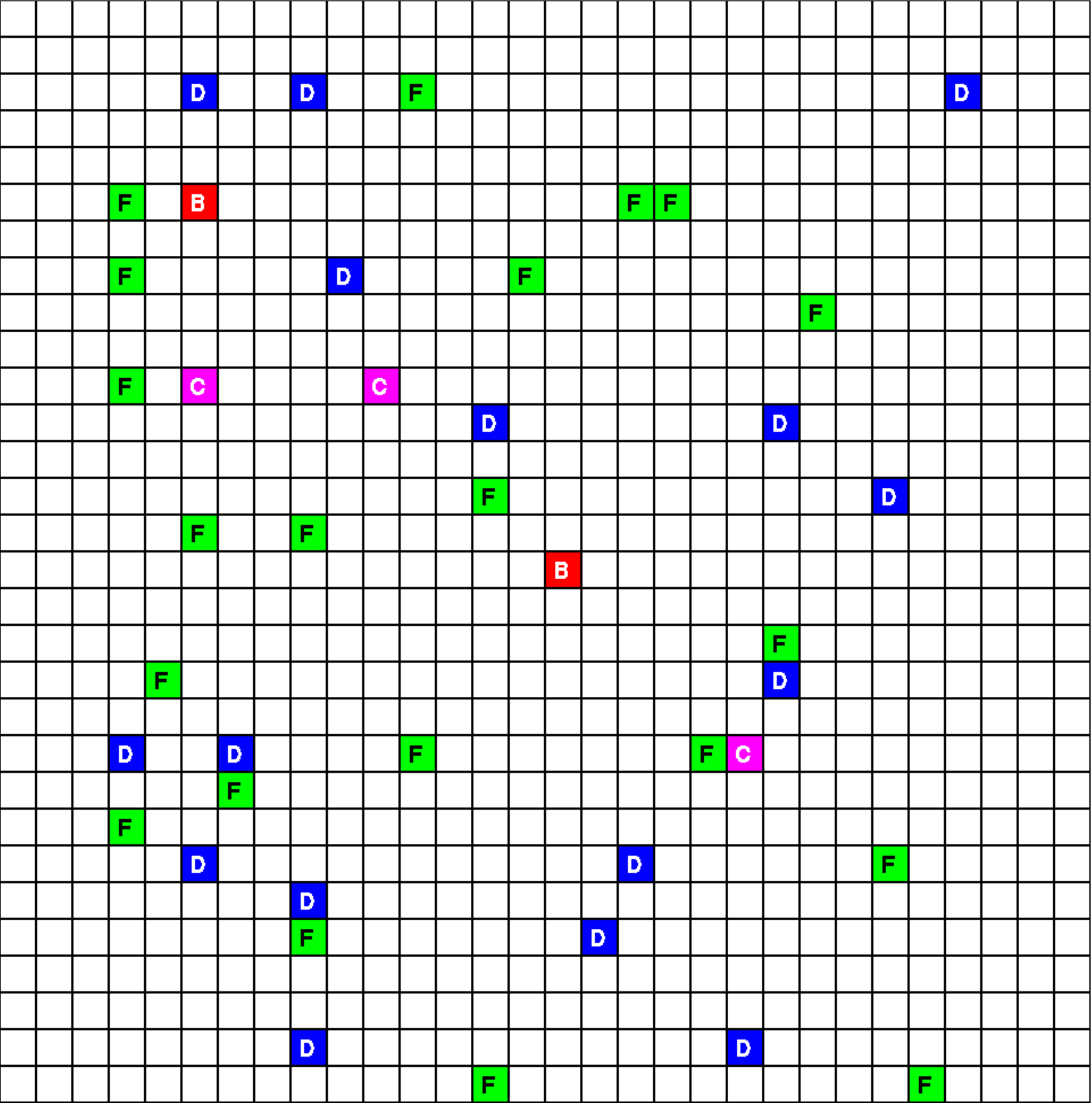}}
	\caption{Illustration of the graphical interface in the deep-warehouse environment using cube-based ASRS configuration.}
	\label{fig:deep_warehouse}
\end{figure}

Figure~\ref{fig:deep_warehouse} illustrates  the state-space in the deep warehouse environment. In a simple cube-based ASRS configuration, the environment consists of (B) passive and (C) active delivery-points, (D) pickup-points, and (F) taxis. Also, the simulator can model other configurations, including advanced cube and shelf-based automated storage and retrieval systems. In the deep warehouse environment, the goal is to store and retrieve goods from one location to another where each cell represents several layers of containers that a taxi can pick up. A taxi (F) receives feedback based on the time used on the task it performs. A taxi can move using a discrete or continuous controller. In discrete mode, the agent can increase and decrease thrust, and move in either direction, including the diagonals. For the continuous mode, all of these actions are floating point numbers between (off) 0 and (on) 1, giving a significantly harder action-space to learn. The simulator also features continuous mode for the state-space, where actions are performed asynchronously to the game loop. It is possible to create custom support modules for mechanisms such as task scheduling, agent controllers and fitness scoring.

A significant benefit of the deep warehouse is that it can accurately model real warehouse environments at high speed. The deep warehouse environment runs 1000 times faster on a single high-end processor core compared to real-world systems measured from the speed improvement by counting how many operations a taxi can do per second. The simulator can be distributed across many processing units to increase the performance further. In our benchmarks, the simulator was able to collect 1 million samples per second during the training of deep learning models using high-performance computing (HPC).

 \section{Experimental Results}
 \label{sec:experiments}
 In this section, we present our preliminary results of applied model-based reinforcement learning using DVAE-2. We aim to answer the following questions. 

\textbf{(1) }Does the DVAE-2 algorithm improve sample efficiency compared to model-free methods? \textbf{(2}) How well do DVAE-2 perform versus model-free methods in the deep warehouse environment? \textbf{(3)} Which of DVAE-2 VRC components is preferred by the model? 
 
 \subsection{The importance of compute}
 According to AI pioneer Richard S. Sutton ``The biggest lesson that can be read from 70 years of AI research is that general methods that leverage computation are ultimately the most effective, and by a large margin.'' \cite{Sutton2019}. It is therefore not surprising that compute is still the most decisive factor when training a large model, also for predictive models. DVAE-2 was initially trained using two NVIDIA 2080 RTX TI GPU cards that, if tuned properly, can operate at approximately 26.9 TFLOPS. For simpler problems, such as grid-warehouses of size \(5 \times 5\) and CartPole, the compute was enough to train the model in 5 minutes, but for larger environments, this time grew exponentially. To somewhat mitigate the computational issue for larger environments, we performed the experiments with approximately 1.25 PFLOPS of compute power. This led to significantly faster training speeds, and made large experiments feasible\footnote{We recognise large experiments to consist of environments where the agents require significant sampling to converge.}

\subsection{Results}
\begin{figure}
	\centering
	\includegraphics[width=0.8\textwidth]{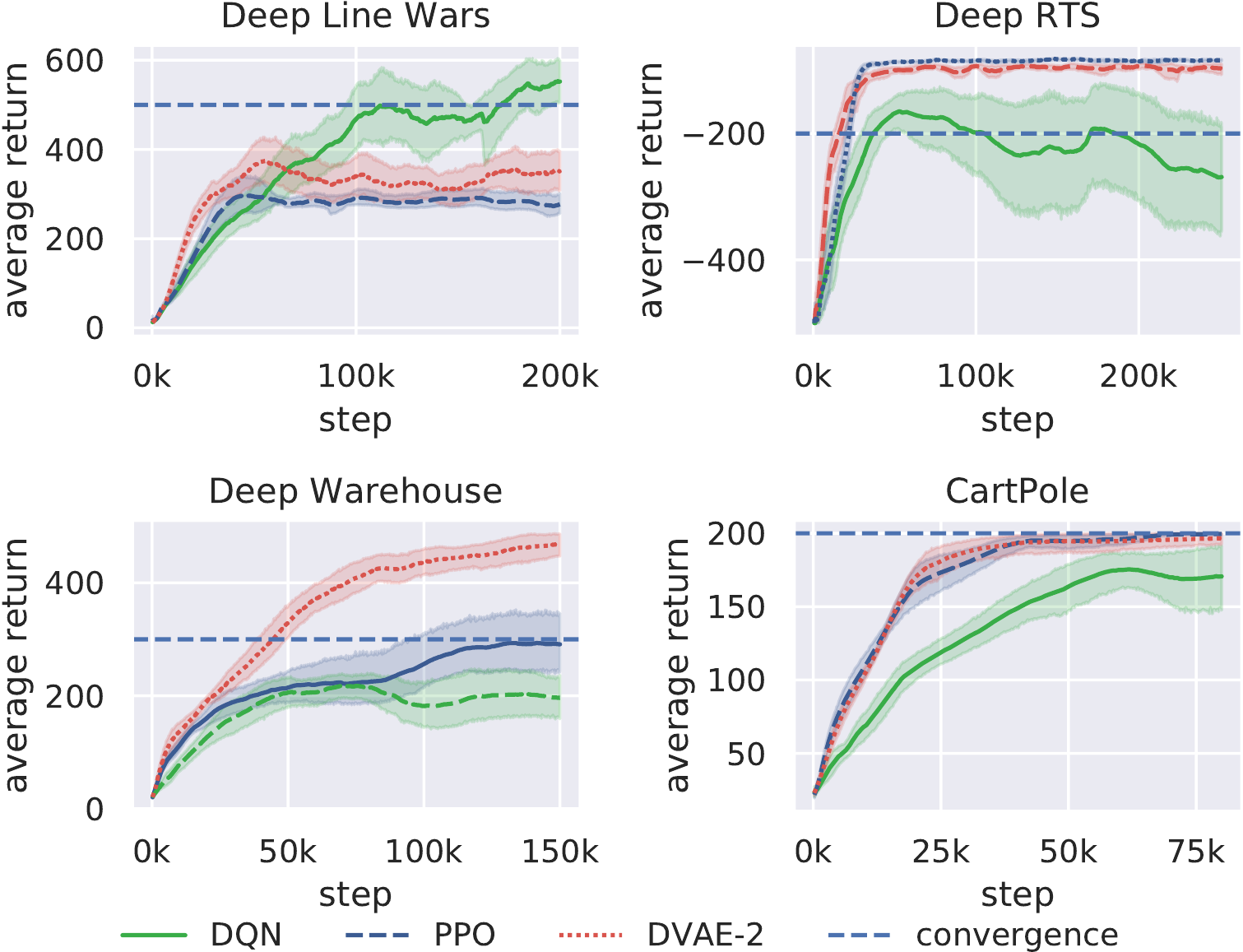}
	
	\caption{We compare DVAE-2 using two baseline algorithms, DQN and PPO. The solid curve illustrates the mean of 12 trials and shaded regions is the standard deviation between all trials. The x-axis shows the number of episodes performed and the y-axis shows the average return.}
	\label{fig:dvae2-results}
\end{figure}

Figure \ref{fig:dvae2-results} shows that the average return value of DVAE-2 training four tasks, including Deep RTS \cite{Andersen2018c}, Deep Warehouse, Deep Line Wars \cite{Andersen2017} and CartPole \cite{Brockman2016a}.

\textbf{Deep Warehouse}: The environment is a contribution in this paper for industry-near testing of autonomous agents. The DVAE-2 algorithm outperforms both PPO and DQN in terms of sampling and performance during 150000 game steps. The score function is a counter of how many tasks the agent has performed during the episode. If the agent manages to collect and retrieve 300 packages, the agent has sufficient performance to beat many handcrafted algorithms in ASRS systems. The environment is multi-agent, and in this experiment, we used a \(30 \times 30\) grid with 20 taxis running the same policy.

\textbf{Deep RTS} is a flexible real-time strategy game (RTS) engine with multiple environments for unit control and resource management. In this experiment, we used the resource harvester environment where the goal is to harvest 500 wood resources before the time limit is up. The score is measured from -500 to 0, where 0 is the best score. For every wood harvested, the score increase with 1. We consider the task mastered if the agent has less than -200 score at the terminal state. DVAE-2 outperform the baseline algorithms in terms of sample efficiency but falls behind PPO in terms of score performance. \cite{Andersen2018c}

\textbf{Deep Line Wars}: Surprisingly, the DQN policy outperforms the DVAE-2 and PPO policy in \( 11 \times 11 \) discrete action-space environment. Because we used PPO as the policy for DVAE-2, we still see a marginal improvement over the same algorithm in a  model-free setting yielding better performance and better sample efficiency. We found that DQN quickly learned the correct Q-values due to the small environment size. In future experiments, we would like to include larger map sizes that would increase the state-space significantly, hence making Q-values more challenging to learn. \cite{Andersen2017}

\textbf{CartPole}: As a simple baseline environment, we use CartPole from the OpenAI Gym environment suite \cite{Brockman2016a}. The goal of this environment is to balance a pole on a moving cart using a discrete action-space of 2 actions. We found that DVAE-2 and PPO had similar performance, but DVAE-2 had marginally better sample efficiency after 25000 steps.

In terms of VRC, the algorithm tended to choose  Convolutional + LSTM and Temporal Convolution and GAN for continuous control tasks (see Figure~\ref{alg:dvae}). It should be noted that PPO and DVAE-2 are presented with the same hyper-parameters, and are therefore directly comparable. We used PPO as our policy for DVAE-2, and we see that DVAE-2 is more sample efficient and performs equally good or better than model-free PPO in all tested scenarios.
 
 \section{Conclusion and Future Work}
 \label{sec:conclusion}
In this paper, we present DVAE-2, a novel model-based reinforcement learning algorithm for improved sample efficiency in environments where sampling is not available. We also present the deep warehouse environment for training reinforcement learning agents in industry-near ASRS systems. This section concludes our work and defines future work for DVAE-2..

Although the deep warehouse does not behave identical to a real-world system, it is adequate to determine the training time and performance. DVAE-2 is presented as a VRC model for training reinforcement learning algorithms with a learned model of the environment. The method is tested in the Deep warehouse several continuous game environments. Our algorithm reduces training time and depends less on data sampled from the real environment compared to model-free methods. 

We find that a carefully tuned policy gradient algorithms can converge to near-optimal behaviour in simulated environments. Model-free algorithms are significantly harder to train in terms of sample efficiency and stability, but perform better if there is unlimited sampling available from the environment.

Our work shows promising results for reinforcement learning agents in ASRS. There are, however, open research questions that are essential for safe deployment in real-world systems. We wish to pursue the following questions to achieve safety deployment in real-world environments. \textbf{(1)} How do we ensure that the agent acts within defined safety boundaries? \textbf{(2)} How would the agent act if parts of the state-space changes to unseen data (i.e. a fire occurs, or a collision between agents.) \textbf{(3)} Can agents with a non-stationary policy function well in a multi-agent setting?

\bibliographystyle{splncs04}
\bibliography{library}

\begin{thebibliography}{10}
\providecommand{\url}[1]{\texttt{#1}}
\providecommand{\urlprefix}{URL }
\providecommand{\doi}[1]{https://doi.org/#1}

\bibitem{Andersen2017}
Andersen, P.A., Goodwin, M., Granmo, O.C.: {Towards a deep reinforcement
  learning approach for tower line wars}. In: Bramer, M., Petridis, M. (eds.)
  Lecture Notes in Computer Science (including subseries Lecture Notes in
  Artificial Intelligence and Lecture Notes in Bioinformatics). vol. 10630
  LNAI, pp. 101--114 (2017). \doi{10.1007/978-3-319-71078-5\_8}

\bibitem{Andersen2018c}
Andersen, P.A., Goodwin, M., Granmo, O.C.: {Deep RTS: A Game Environment for
  Deep Reinforcement Learning in Real-Time Strategy Games}. Proceedings of the
  IEEE International Conference on Computational Intelligence and Games  (aug
  2018), \url{http://arxiv.org/abs/1808.05032}

\bibitem{Andersen2018b}
Andersen, P.A., Goodwin, M., Granmo, O.C.: {The Dreaming Variational
  Autoencoder for Reinforcement Learning Environments}. In: {Max Bramer},
  Petridis, M. (eds.) Artificial Intelligence, vol. 11311, pp. 143--155.
  Springer, Cham, xxxv edn. (dec 2018). \doi{10.1007/978-3-030-04191-5\_11},
  \url{http://link.springer.com/10.1007/978-3-030-04191-5{\_}11}

\bibitem{Azar2019}
Azar, M.G., Piot, B., Pires, B.A., Grill, J.B., Altch{\'{e}}, F., Munos, R.:
  {World Discovery Models}. arxiv preprint arXiv:1902.07685  (feb 2019),
  \url{http://arxiv.org/abs/1902.07685}

\bibitem{Blundell2016}
Blundell, C., Uria, B., Pritzel, A., Li, Y., Ruderman, A., Leibo, J.Z., Rae,
  J., Wierstra, D., Hassabis, D.: {Model-Free Episodic Control}. arxiv preprint
  arXiv:1606.04460  (jun 2016), \url{http://arxiv.org/abs/1606.04460}

\bibitem{Botvinick2019}
Botvinick, M., Ritter, S., Wang, J.X., Kurth-Nelson, Z., Blundell, C.,
  Hassabis, D.: {Reinforcement Learning, Fast and Slow.} Trends in cognitive
  sciences  \textbf{23}(5),  408--422 (may 2019).
  \doi{10.1016/j.tics.2019.02.006},
  \url{http://www.ncbi.nlm.nih.gov/pubmed/31003893}

\bibitem{Brockman2016a}
Brockman, G., Cheung, V., Pettersson, L., Schneider, J., Schulman, J., Tang,
  J., Zaremba, W.: {OpenAI Gym}. arxiv preprint arXiv:1606.01540  (jun 2016),
  \url{http://arxiv.org/abs/1606.01540}

\bibitem{Buckman2018}
Buckman, J., Hafner, D., Tucker, G., Brevdo, E., Lee, H.: {Sample-Efficient
  Reinforcement Learning with Stochastic Ensemble Value Expansion}. Advances in
  Neural Information Processing Systems 32 pp. 8224--8234 (jul 2018),
  \url{http://arxiv.org/abs/1807.01675}

\bibitem{Chua2018}
Chua, K., Calandra, R., McAllister, R., Levine, S.: {Deep Reinforcement
  Learning in a Handful of Trials using Probabilistic Dynamics Models}.
  Advances in Neural Information Processing Systems 31  (may 2018),
  \url{http://arxiv.org/abs/1805.12114}

\bibitem{greensmith2004variance}
Greensmith, E., Bartlett, P.L., Baxter, J.: {Variance reduction techniques for
  gradient estimates in reinforcement learning}. Journal of Machine Learning
  Research  \textbf{5}(Nov),  1471--1530 (2004)

\bibitem{Gregor2019}
Gregor, K., Rezende, D.J., Besse, F., Wu, Y., Merzic, H., van~den Oord, A.:
  {Shaping Belief States with Generative Environment Models for RL}. arxiv
  preprint arXiv:1906.09237  (jun 2019), \url{http://arxiv.org/abs/1906.09237}

\bibitem{Ha2018a}
Ha, D., Schmidhuber, J.: {Recurrent World Models Facilitate Policy Evolution}.
  Advances in Neural Information Processing Systems 31  (sep 2018),
  \url{http://arxiv.org/abs/1809.01999}

\bibitem{Ha2018}
Ha, D., Schmidhuber, J.: {World Models}. arxiv preprint arXiv:1803.10122  (mar
  2018). \doi{10.5281/zenodo.1207631}, \url{https://arxiv.org/abs/1803.10122}

\bibitem{Hafner2018}
Hafner, D., Lillicrap, T., Fischer, I., Villegas, R., Ha, D., Lee, H.,
  Davidson, J.: {Learning Latent Dynamics for Planning from Pixels}.
  Proceedings of the 36 th International Conference on Machine Learning  (nov
  2018), \url{http://arxiv.org/abs/1811.04551}

\bibitem{Higgins2016}
Higgins, I., Matthey, L., Pal, A., Burgess, C., Glorot, X., Botvinick, M.,
  Mohamed, S., Lerchner, A.: {beta-VAE: Learning Basic Visual Concepts with a
  Constrained Variational Framework}. International Conference on Learning
  Representations  (nov 2016), \url{https://openreview.net/forum?id=Sy2fzU9gl}

\bibitem{Janner2019}
Janner, M., Fu, J., Zhang, M., Levine, S.: {When to Trust Your Model:
  Model-Based Policy Optimization}. arXiv preprint arXiv:1906.08253  (jun
  2019), \url{http://arxiv.org/abs/1906.08253}

\bibitem{Kaelbling1996}
Kaelbling, L.P., Littman, M.L., Moore, A.W.: {Reinforcement Learning: A
  Survey}. Journal of Artificial Intelligence Research  (apr 1996).
  \doi{10.1.1.68.466}, \url{http://arxiv.org/abs/cs/9605103}

\bibitem{Liang2018}
Liang, X., Wang, Q., Feng, Y., Liu, Z., Huang, J.: {VMAV-C: A Deep
  Attention-based Reinforcement Learning Algorithm for Model-based Control}.
  arxiv preprint arXiv:1812.09968  (dec 2018),
  \url{http://arxiv.org/abs/1812.09968}

\bibitem{Mnih2013}
Mnih, V., Kavukcuoglu, K., Silver, D., Graves, A., Antonoglou, I., Wierstra,
  D., Riedmiller, M.: {Playing Atari with Deep Reinforcement Learning}. Neural
  Information Processing Systems  (dec 2013),
  \url{http://arxiv.org/abs/1312.5602}

\bibitem{Roodbergen2009}
Roodbergen, K.J., Vis, I.F.A.: {A survey of literature on automated storage and
  retrieval systems}. European Journal of Operational Research  (2009).
  \doi{10.1016/j.ejor.2008.01.038}

\bibitem{Schulman2017}
Schulman, J., Wolski, F., Dhariwal, P., Radford, A., Klimov, O.: {Proximal
  Policy Optimization Algorithms}. arxiv preprint arXiv:1707.06347  (jul 2017),
  \url{http://arxiv.org/abs/1707.06347}

\bibitem{Sutton2019}
Sutton, R.S.: {The Bitter Lesson} (2019),
  \url{http://www.incompleteideas.net/IncIdeas/BitterLesson.html}

\bibitem{Sutton2018}
Sutton, R.S., Barto, A.G.: {Reinforcement learning: An introduction}. MIT Press
  (2018)

\end{thebibliography}

 \end{document}